%% file: main.tex
\documentclass[10pt,twocolumn,letterpaper]{article}

\usepackage{cvpr}              
\usepackage{float}

\usepackage{amsmath} 
\usepackage[normalem]{ulem}
\useunder{\uline}{\ul}{}
\usepackage{makecell}
\usepackage{graphicx}  
\usepackage{float}  
\usepackage{tabularx}
\usepackage{arydshln}
\usepackage{multirow}
\usepackage{placeins}
\usepackage{stfloats}

\input{preamble}
\definecolor{cvprblue}{rgb}{0.21,0.49,0.74}
\usepackage[pagebackref,breaklinks,colorlinks,allcolors=cvprblue]{hyperref}

\def\paperID{10873} 
\def\confName{CVPR}
\def\confYear{2026}

\title{FSFSplatter: Geometrically Accurate Reconstruction with Free Sparse-view Images within 2 minutes}

 \author{Yibin Zhao$^{1}$ \qquad Yihan Pan$^{1}$ \qquad Jun Nan$^{1}$ \qquad Liwei Chen$^{2}$ \qquad Jianjun Yi$^{1}$ \\
 $^{1}$East China University of Science and Technology, Shanghai\\
 $^{2}$Shanghai Xiaoyuan Innovation Center, Shanghai\\
 {\tt\small  \url{https://github.com/Zhaoyibinn/FSFSplatter}}\\
 }

\begin{document}

\maketitle




\input{sec/0_abstract}

\input{sec/1_intro}

\input{sec/2_relatedworks}

\input{sec/3_method}
\input{sec/4_exp}

\input{sec/5_conclusion}
\input{sec/6_Acknowledgments}


{
    \small
    \bibliographystyle{ieeenat_fullname}
    \bibliography{main}
}

\end{document}

%% file: sec/0_abstract.tex
\begin{abstract}
Gaussian Splatting has become a leading reconstruction technique, known for its high-quality novel view synthesis and detailed reconstruction. However, most existing methods require dense, calibrated views. Reconstruction from free sparse-view images often leads to poor surface due to limited overlap and overfitting.
We introduce FSFSplatter for \textbf{f}ast geometrically accurate reconstruction from \textbf{f}ree \textbf{s}parse-view images. 
Our method integrates end-to-end dense Gaussian scene initialization and geometry-enhanced scene optimization.
Specifically, FSFSplatter employs a large transformer to encode multi-view images and generates a dense and geometrically consistent Gaussian scene initialization via a batch based self-splitting Gaussian head. It eliminates local floaters through contribution-based pruning and mitigates overfitting by leveraging depth and multi-view feature supervision, along with differentiable camera parameters within 2 minutes.
FSFSplatter outperforms current state-of-the-art methods on widely used DTU, Replica, and BlendedMVS datasets.
\vspace{-4mm}
\end{abstract}

%% file: sec/1_intro.tex
%

\section{Introduction}

\label{sec:intro}

\hspace{1em}Surface reconstruction and novel view synthesis (NVS) from free multi-view RGB images are long-standing tasks in autonomous driving \cite{geiger2012areweready}, computer graphics \cite{2dgs_Huang}, and virtual reality \cite{luo2020consistentvideo}. As typical methods require dense, calibrated images, accurate reconstruction and NVS from free sparse-view inputs remain particularly challenging, yet crucial for scenes with only limited RGB captures.

Traditional approaches often decompose the problem into separate stages of camera pose estimation, dense reconstruction, and surface extraction, typically by converting dense point clouds into surfaces via signed distance functions (SDFs). This pipeline, however, introduces errors at each stage, leading to irreversible error accumulation.

Recent studies have explored end-to-end models for free image reconstruction. For example, DUSt3R \cite{dust3rwang} learns a robust image-to-pointmap mapping from image pairs, with several variants \cite{Yang_2025_Fast3R,mast3r_eccv24}. Nevertheless, its network architecture is incapable of processing multiple views in a single forward pass, resulting in accumulated inconsistencies. VGGT \cite{wang2025vggt} employs alternating global and local attention mechanisms, enabling 3D reconstruction based on point clouds from an arbitrary number of free images.

The sparsity of point clouds, however, makes them a suboptimal representation for many downstream applications (e.g., NVS). Recent breakthroughs in differentiable rendering \cite{kerbl3DGS,mildenhall2020nerf} have demonstrated unprecedented quality in multi-view reconstruction. However, most of these methods struggle to generalize to free sparse-view views, as insufficient image overlap often leads to the failure of Structure-from-Motion (SfM), and the inherent ambiguity in dense reconstruction arises from view extrapolation and occlusion under sparse views.
Methods like PF-LRM \cite{PF-LRM,jiang2022LEAP} employ a transformer to map RGB images into a NeRF representation. Nevertheless, the surface bias of NeRF \cite{wang2021neus} and the separate point cloud head for camera pose estimation limit the final reconstruction quality. In contrast, Gaussian Splatting (GS), due to the explicit scene representation and unbiased geometry, has shown advantages in surface reconstruction \cite{guedon2023sugar,2dgs_Huang}. FreeSplatter \cite{xu2024freesplatter} constructs a GS scene in a feed-forward manner from a single, unconstrained RGB image, and subsequently estimates camera parameters and aligns multiple views using an optimizer, which introduces additional errors.

To address point clouds' sparsity and additional errors from post-processing, we employ a large transformer as the backbone network, which regresses scale-consistent camera parameters, depth maps, and a Gaussian scene via independent prediction heads. To predict dense and geometrically accurate Gaussian primitives, we specifically design a geometry and global feature based Gaussian self-splitting head. A semi-dense Gaussian scene is first obtained by fusing the pixel-aligned depth map and Dense Prediction Transformers (DPT)\cite{Ranftl2021DPT} decoded features. Subsequently, an Encoder-Decoder architecture is utilized to compute offset maps and perform Gaussian self-splitting, yielding the dense Gaussian scene. To enhance the overall geometric stability and eliminate local floaters, we apply a contribution-based pruning strategy to obtain the initial scene. Furthermore, to address the issue of local view overfitting during the optimization process under sparse inputs, we propose a geometry-enhanced scene optimization supervised by monocular depth and multi-view features, which significantly improves both surface reconstruction and NVS quality within a short time.

Our main contributions can be summarized as follows:
\begin{enumerate}
    \item We propose \textbf{FSFSplatter}, as shown in \cref{toutu}, a fast geometrically accurate reconstruction algorithm from free sparse-view images. Compared to prior works, FSFSplatter achieves accurate reconstruction results both with and without per-scene optimization.
    
    \item The Gaussian head performs Gaussian self-splitting based on high-dimensional point clouds, coupled with contribution-based pruning, enabling precise initialization of the dense Gaussian scene. Furthermore, we introduce a geometry-enhanced scene optimization supervised by monocular depth and multi-view features, which effectively addresses the common issue of local view overfitting in sparse-view optimization.
    
    \item We demonstrate state-of-the-art (SOTA) performance on widely-used DTU, Replica, and BlendedMVS datasets upon surface reconstruction and NVS. 
\end{enumerate}

%% file: sec/2_relatedworks.tex
\section{Related Works}
\subsection{Surface reconstruction}

\hspace{1em}Surface reconstruction from RGB images is a fundamental problem in computer vision. Traditional methods typically represent scenes as point clouds \cite{schonberger2016colmap}, depth maps \cite{galliani2015massively}, or voxels \cite{kostrikov2014probabilistic}. In recent years, with the advancement of computational hardware and neural network technologies, implicit scene representations have been widely adopted. NeRF \cite{mildenhall2020nerf} has been applied to NVS and 3D reconstruction, undergoing rapid development. 
3D Gaussian Splatting (3DGS) represents a scene using a set of explicit Gaussian primitives \cite{kerbl3DGS}, enabling fast and high-quality NVS and reconstruction. However, 3DGS lacks expressive capacity for scene geometry. Subsequent work has achieved accurate surface reconstruction by constraining the shapes of Gaussian primitives to align with the surface \cite{2dgs_Huang,guedon2023sugar}. Through depth map rendering and fusion, fast and accurate geometric surface reconstruction can be realized. Furthermore, the reconstruction quality can be further enhanced by mapping algorithms from colored point clouds to Gaussians \cite{wang2024pfgs,gaupcrender}.

\subsection{Free sparse-view reconstruction}
\hspace{1em}However, most of the aforementioned methods require dense camera views with known parameters. Under sparse views, Structure-from-Motion (SfM) often fails due to insufficient image overlap. Even with constrained camera parameters, errors in the camera parameters and sparsity can gradually lead to overfitting to individual views, resulting in erroneous geometric surfaces.

In traditional pipelines\cite{schonberger2016colmap, schoenberger2016colmapmvs}, image-based 3D reconstruction involves separate stages such as feature extraction, feature matching, bundle adjustment (BA), and plane sweeping. 
However, both camera pose estimation from sparse views and NVS-based dense reconstruction methods suffer from unavoidable ambiguities caused by extrapolation and occlusion \cite{zhang2020nerf++}. 
Some approaches address this by leveraging large-scale dataset pre-training for better generalization to other datasets, but often require several days of training \cite{zhang2020nerf++}. Methods focusing on per-scene optimization introduce additional supervision to enhance geometric stability during the optimization process \cite{huang2025fatesgs, xiong2023sparsegs}. 

\begin{figure*}[!t]
\centering
\includegraphics[width = 0.99\textwidth]{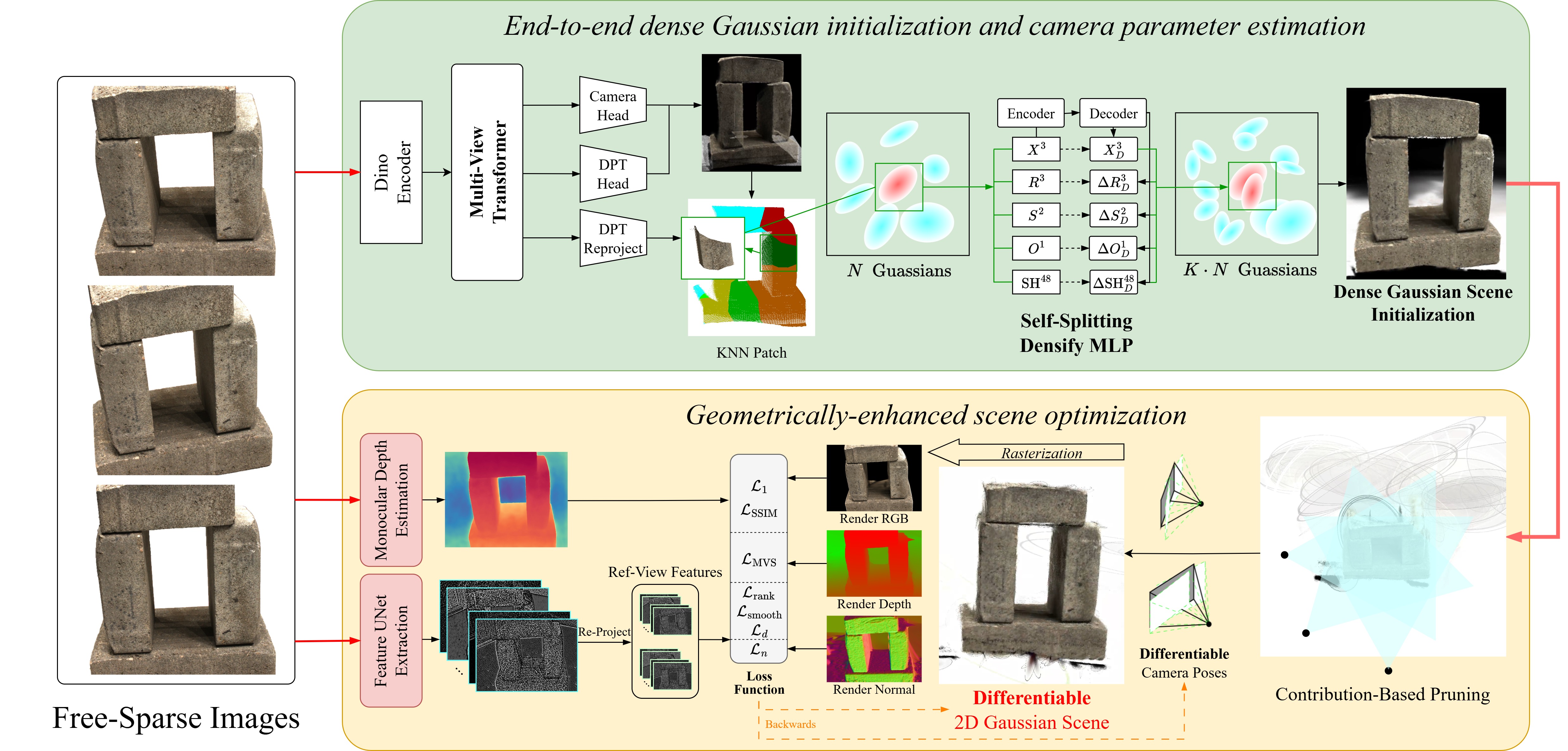}
\caption{\textbf{Overview of FSFSplatter}. We employ a large multi-view transformer as the backbone and generate high-dimensional semi-dense Gaussian scenes through independent heads. These are subsequently mapped into a dense Gaussian scene via a patch-based self-splitting densify MLP, forming an end-to-end framework for dense Gaussian initialization and camera parameter estimation. After differentiable Gaussian scene construction from initialization and contribution-based pruning with differentiable camera parameters, the reconstruction achieves geometric enhancement by joint supervision of monocular depth, multi-view stereo features, and RGBs during the optimization, effectively mitigating error surfaces caused by overfitting to free sparse-view views.}\label{overview}
\end{figure*}

With the growing recognition of the potential of large Transformer models, pose estimation and 3D reconstruction are increasingly being implemented in an end-to-end pipeline. Dust3R \cite{dust3rwang} and its subsequent developments \cite{Yang_2025_Fast3R, mast3r_eccv24} achieve accurate camera parameter estimation and PointMap prediction from two images. VGGT facilitates joint camera parameter estimation and 3D reconstruction from multiple images in a single forward pass by leveraging alternating global and local attention \cite{wang2025vggt}.
FreeSplatter \cite{xu2024freesplatter} maps free, single views to a Gaussian scene and performs multi-view camera pose estimation and scene alignment based on optimization, treating scene generation and camera pose estimation as independent steps.

%% file: sec/3_method.tex
\section{Method}

\hspace{1em}We propose a fast reconstruction method based on free sparse-view RGB images that enables robust geometrically accurate surface and NVS from only 3 input images within 2 minutes.

Our framework consists of two main components: an end-to-end differentiable dense Gaussian initialization and camera parameter estimation, and a geometry-enhanced Gaussian scene optimization.

\subsection{Preliminaries}
\subsubsection{Gaussian Splatting}
\hspace{1em}2DGS flattens 3D Gaussians into 2D primitives \cite{2dgs_Huang}, enabling precise surface reconstruction by closely adhering the Gaussians to the actual surfaces. Each Gaussian primitive is defined by its position $\mu^3$, rotation $R^4$, scale $S^2$, opacity $O^1$, and spherical harmonics (SH) coefficients $C^{48}$ for representing color.
The splatting of Gaussian primitives from the Gaussian coordinate system to the pixel coordinate system is achieved via a homogeneous transformation matrix $\mathbf{H}$, as shown in \cref{2dgs_splat}.

\begin{equation}\label{2dgs_splat}
\begin{aligned} 
\mathbf{u}&=\mathbf{W} \mathbf{H}(s_1, s_2, 1,1)^{\top} 
\end{aligned}
\end{equation}

Multiple Gaussian primitives are composited via $\alpha$-blending, as shown in \cref{2dgs_a-blend}. Here, $\mathcal{G}$ denotes the set of primitives influencing the pixel, and $\alpha$ is computed based on the Gaussian's covariance matrix $\boldsymbol{\Sigma}$, pixel coordinates $\boldsymbol{p}$, and the projected Gaussian center $\boldsymbol{\mu}$.
\begin{equation}\label{2dgs_a-blend}
\begin{aligned} 
\mathbf{C} &= \sum_{ i \in \mathcal{G}} \mathbf{c}_i \alpha_i \prod_{i-1}^{n}(1- \alpha_n ) \ \  where \\
 \alpha_n &= \sigma_{n} \cdot \exp \left(-\frac{1}{2}\left(\boldsymbol{p}-\boldsymbol{\mu}_{n}\right)^{T} \boldsymbol{\Sigma}_{n}^{-1}\left(\boldsymbol{p}-\boldsymbol{\mu}_{n}\right)\right)
\end{aligned}
\end{equation}

\subsubsection{Visual Geometry Grounded Transformer}
\hspace{1em}To address the issue of error accumulation caused by multiple stages in traditional 3D reconstruction pipelines, VGGT \cite{wang2025vggt} employs a large feed-forward neural network to directly infer all key attributes from multiple input views. These attributes include camera parameters $\mathbf{g}_{i}$, depth maps $D_{i}$, pointmaps $P_{i}$, and point trajectories $T_{i}$. The backbone network incorporates alternating attention mechanisms. Utilizing a shared backbone with separate, dedicated prediction heads enables the simultaneous estimation of all required 3D attributes.


Our core contribution lies in two parts: End-to-end initialization (Sec. \ref{sec_gsinit}) and geometry enhanced optimization (Sec. \ref{sec_gsopt}). They collectively addressing the catastrophic geometric degradation encountered in sparse-view reconstruction: the initialization is essential for the optimization to converge on geometry, and the optimization strategy is essential for refining the initialization without overfitting.
\subsection{End-to-end dense Gaussian initialization and camera parameter estimation}
\label{sec_gsinit}

\subsubsection{Network architecture}
\hspace{1em}Inspired by recent advances in 3D depth estimation \cite{wang2025vggt, dust3rwang}, we adopt a large transformer as the backbone for our end-to-end dense Gaussian initialization and camera parameter estimation framework. Images are first encoded into tokens $(T_i)^3$ using DINOv2 \cite{oquab2023dinov2}. These tokens are then processed through the main network structure, which employs a large multi-view Transformer.


\begin{figure}[htpb]
\centering
\includegraphics[width = 0.49\textwidth]{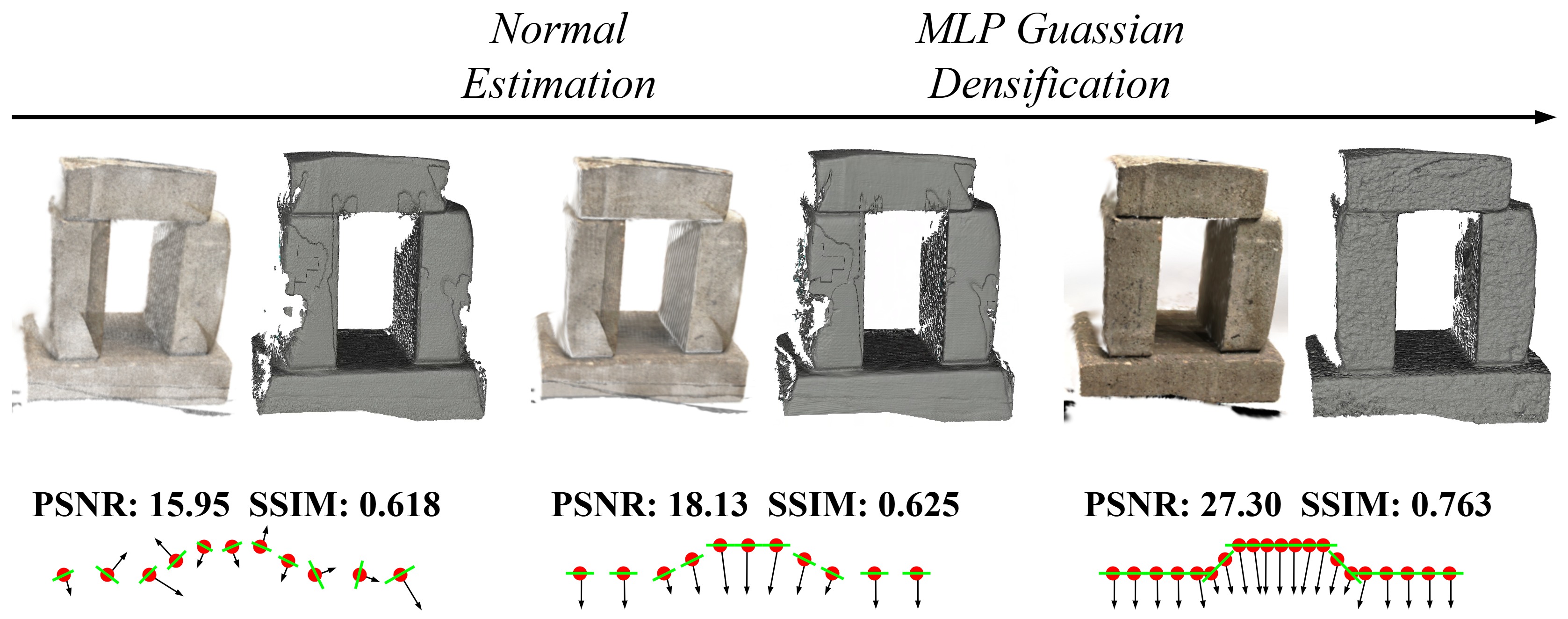}
\caption{
By incorporating explicit normal estimation and an MLP-based self-splitting Gaussian densification, the quality of both NVS and surface reconstruction is significantly improved.}\label{vis_gsinit_method}
\end{figure}

Based on the outputs from the camera head and the DPT head, we can rapidly perform reprojection to obtain pixel-aligned 3D point clouds. Concurrently, we design an independent DPT reprojection module that utilizes positional encoding to reproject the encoded tokens into pixel-aligned high-dimensional features $\mathcal{F}_P$.
We employ an encoder-decoder architecture $D$ to map the aforementioned information into dense Gaussian primitive attributes. Specifically, PointMLP \cite{pointmlpma} is adopted to encode the geometric information of the point cloud, yielding high-dimensional geometric features $\mathcal{F}_c$. Subsequently, $\mathcal{F}_P$ is concatenated channel-wise with explicit Gaussian attributes—including $\text{SH}^{48}_P$, $R^4_P$, $S^2_P$, and $\mathcal{F}_P$—to form the encoded representation.
We then design mutually independent decoders for the attribute variations $\Delta \mathcal{G}$ of all Gaussian primitives. These decoders predict the densification tendency of the original semi-dense Gaussian scene, as shown in \cref{overview} and \cref{vis_gsinit_method}. Each Gaussian primitive splits into $N$ new primitives based on its predicted variation, as formulated in \cref{GSinit_decoder}.

\begin{equation}\label{GSinit_decoder}
\begin{aligned} 
\Delta \mathcal{G}_{N\cdot P} &= \{\Delta X^3_{N\cdot P}, \Delta R^4_{N\cdot P},\Delta S^2_{N\cdot P},\Delta O^1_{N\cdot P},\Delta \text{SH}^{48}_{N\cdot P} \} \\
&= D(\text{Cat}(\mathcal{F}_P,\text{SH}^{48}_P,R^4_P,S^2_P,\mathcal{F}_c))\\
\mathbb{D}(&\mathcal{G}_P) = \mathcal{G}_P + \Delta \mathcal{G}_{N\cdot P}
\end{aligned}
\end{equation}

Drawing on advanced point cloud processing techniques \cite{shapenetMarcin,gaupcrender}, we adopt point cloud patches as the input units to our network to enhance the generalization capability across diverse scenes.
We employ K-Nearest Neighbors (KNN) to partition the high-dimensional point cloud into randomly sampled local patches. Each patch is processed through an encoder-decoder architecture to generate densified Gaussian primitives. Finally, after de-normalization, the processed batches are merged to form the complete Gaussian scene, as expressed in \cref{gsinit_batch}. Here, $\mathbf{N}$ and $\mathbf{N}'$ denote the normalization and de-normalization operations; $\mathbb{B}$ and $\mathbb{B}'$ represent batch partitioning and merging; and $\mathbb{D}$ signifies the aforementioned densification process.

\begin{equation}\label{gsinit_batch}
\begin{aligned} 
\mathcal{G} = \mathbf{N}^{\prime} \biggl\{ \mathbb{B}^{\prime} \Bigl[ \mathbb{B}\bigl( \mathbf{N}(\mathcal{G}_{init}) \bigr) + \mathbb{D}\bigl( \mathbb{B}(\mathbf{N}(\mathcal{G}_{init})) \bigr) \Bigr] \biggr\}
\end{aligned}
\end{equation}

\subsubsection{Training strategy}
\hspace{1em}Our entire pipeline is fully end-to-end differentiable, enabling comprehensive training based on $\mathcal{L}_\text{RGB-D}$ and $\mathcal{L}_\text{Cam}$. Here, $\mathcal{L}_\text{RGB-D}$ comprises an RGB-based loss combining $\mathcal{L}_1$ and $\mathcal{L}_\text{SSIM}$, along with a depth-based loss $\mathcal{L}_\text{D}$. The camera parameters are defined by the focals $\mathbf{F}\in \mathbb{R}^2$ and the extrinsic matrix $\mathbf{T} = [ \mathbf{t} \in \mathbb{R}^3, \mathbf{q} \in \mathbb{R}^4]$, forming the camera loss $\mathcal{L}_\text{Cam} = \mathcal{L}_\mathbf{F} + \mathcal{L}_\mathbf{T}$.

A straightforward approach involves training all parameters of the entire pipeline directly. However, in practice, since both the input and the final pipeline output are RGB images, this often leads to overfitting on the NVS of local views during training.
To mitigate this, we adopt a geometry-progressive training strategy.
We initialize our network backbone with pre-trained weights from VGGT-1B \cite{wang2025vggt}.
During this stage, the losses $\mathcal{L}_\text{D}$ and $\mathcal{L}_\text{Cam}$ are enabled to ensure geometric stability, maintaining consistency with the training methodology of VGGT.

To achieve a high-quality initialization of the Gaussian scene and prevent overfitting, we subsequently freeze the DPT Head, Camera Head, and the backbone network. This freezing preserves geometric stability. The DPT reproject Head and the densification MLP are then trained independently. In this stage, KNN Patch partitioning is disabled, allowing for global densification and estimation across the entire scene, which serves as a form of pre-training. The losses $\mathcal{L}_1$ and $\mathcal{L}_\text{SSIM}$ are enabled during this stage.

Finally, we unfreeze all parameters of the pipeline, enabling the backbone network to learn the mapping from RGB images to Gaussian attributes directly. The KNN Patch operation is reinstated during the densification process to enhance generalization. The complete set of loss functions is activated throughout this final training stage.


\subsection{Geometry enhanced Gaussian scene optimization}
\label{sec_gsopt}
\hspace{1em}Due to free sparse-view images, direct optimization using RGB-based losses introduces inevitable multi-view ambiguity \cite{zhang2020nerf++}. This often leads to color overfitting on sparse views and results in catastrophic geometry degradation. To address this, we propose a geometry-enhanced optimization strategy for Gaussian scenes, as shown in \cref{overview}.
\vspace{-5mm}
\paragraph{Differentiable Gaussian scene construction}


While sub-pixel level Gaussian scene initialization provides a strong prior, it also introduces numerous occluded or invisible primitives, which can not receive gradient updates through backpropagation and be pruned simply by filtering low-opacity elements \cite{2dgs_Huang,kerbl3DGS}.
Therefore, we perform contribution-based Gaussian pruning before optimization.

\begin{figure}[htpb]
\centering
\includegraphics[width = 0.45\textwidth]{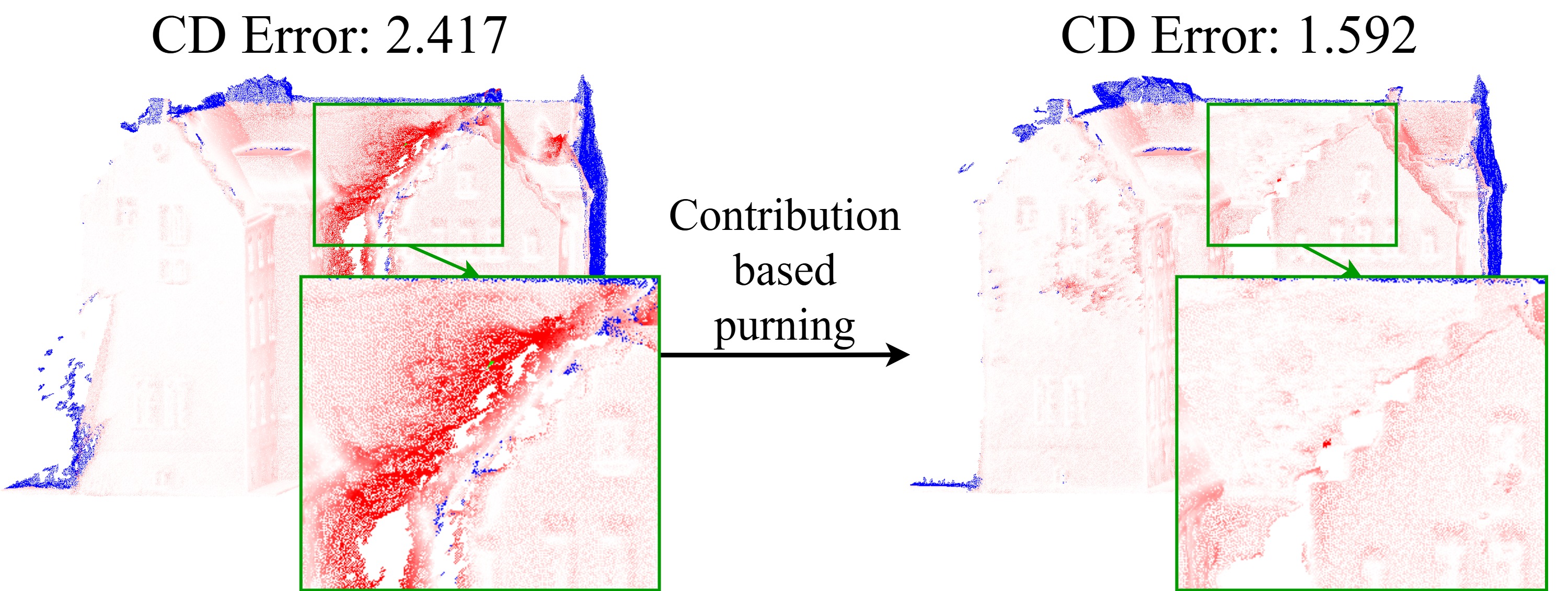}
\caption{Contribution-based Gaussian pruning for scene initialization}\label{fig_trim_vis}
\end{figure}

Referring to \cref{2dgs_a-blend}, the contribution $\mathbf{C}_n$ of a Gaussian primitive $\mathcal{G}_n$ can be quantified by its weight in $\alpha$-blending process: $ \alpha_i \prod_{i-1}^{n}(1- \alpha_n )$, as shown in \cref{trim_con}. Here, $\mathcal{P}_n$ denotes pixels influenced by $\mathcal{G}_n$. To prevent unfair large contributions from Gaussian scales, we normalize the contribution by the number of occupied pixels $|\mathcal{P}_n|$. After computing contributions for all primitives, we sort them and prune those with low contribution or low opacity, as shown in \cref{fig_trim_vis}.

\begin{equation}\label{trim_con}
\begin{aligned} 
\mathbf{C}_n = \sum_{k = 1}^{3}\frac{1}{|\mathcal{P}_n|}  \sum_{p \in \mathcal{P}_n }(\alpha_i^p \prod_{i-1}^{n}(1- \alpha_n^p )) 
\end{aligned}
\end{equation}

GS's original rasterizer does not support the backward propagation of camera parameters, which is detrimental for pose-free 3D reconstruction, as it tends to amplify subtle multi-view inconsistencies during optimization.
Following related works \cite{keetha2024splatam}, we consistently position camera poses at the origin point.
Before rasterization, we apply inverse transformations of camera poses $\mathbf{T}^\text{cam}_k$ to all Gaussian primitives, thereby achieving the same NVS as in the original framework. Throughout this process, $\mathbf{T}^\text{cam}_k$ remains differentiable. Any loss based on rasterized images will propagate back to camera poses before being passed to Gaussian primitives during backpropagation.
Owing to the sparsity of viewpoints, our approach enables rapid and mutually independent pose optimization across different views.

\vspace{-3mm}

\paragraph{Optimization with geometry-based prior}

Following prior sparse-view reconstruction methods \cite{huang2025fatesgs, wu2025sparse2dgs}, our loss function combines RGB terms ($\mathcal{L}_1$, $\mathcal{L}_\text{SSIM}$), a normal term ($\mathcal{L}_n$), depth-related terms ($\mathcal{L}_\text{rank}$, $\mathcal{L}_\text{smooth}$, $\mathcal{L}_d$), and a multi-view feature term ($\mathcal{L}_\text{MVS}$). The formulations of $\mathcal{L}_1$, $\mathcal{L}_\text{SSIM}$, $\mathcal{L}_d$, and $\mathcal{L}_n$ follow 2DGS \cite{2dgs_Huang} and 3DGS \cite{kerbl3DGS}.

To mitigate scale ambiguity, we supervise rasterized depth $D_{\text{re}}$ using monocular depth estimates $D_{\text{est}}$ \cite{ke2025marigold} via a ranking loss (\cref{rank_loss}). Pixel pairs $(p_1,p_2)$ are randomly sampled from local patches $\mathcal{P}$ through strategy $\mathcal{R}$. Here, $\text{sgn}$ is the sign function, $m$ is a margin, and $\sigma$ denotes ReLU.
\begin{equation}\label{rank_loss}
\begin{aligned} 
\mathcal{L}_{\text{rank}} =& \sum_{ \mathcal{P}}\sigma \biggl(\operatorname{sgn}\big( \mathcal{S}(D_{\text{pre}},p_1,p_2)\cdot\mathcal{S}(D_{\text{re}},p_1,p_2)\big)+m \biggr)  \\
&where \ \ \ \mathcal{S}(D,p_1,p_2) = D(p_1) - D(p_2)
\end{aligned}
\end{equation}
Concurrently, we enforce a depth smoothness loss $\mathcal{L}_\text{smooth}$. 
As shown in \cref{smooth_loss}, $n_1$ and $n_2$ are small thresholds used to identify edges in $D_{\text{est}}$. 
\begin{equation}\label{smooth_loss}
\begin{aligned} 
\mathcal{L}_\text{smooth} = \sum_{p_1,p_2 \in \mathcal{P}} 
&\text{sgn}\big( |D_\text{pre}(p_1) - D_\text{pre}(p_2)| - n_1 \big) \\
&\cdot \sigma\big( |D_\text{re}(p_1) - D_\text{re}(p_2)| - n_2 \big)
\end{aligned}
\end{equation}
To address inconsistencies in multi-view illumination \cite{2018UnsupervisedMonocularDepth}, we employ a multi-view feature alignment loss.
Specifically, high-dimensional features are extracted using a pre-trained U-Net $\mathbf{U}$ \cite{zhang2020visibility}, reprojected to reference views via $\mathcal{R}_r$, and compared as defined in \cref{fea_loss}.
\begin{equation}\label{fea_loss}
\begin{aligned} 
\mathcal{L}_\text{MVS} = \sum_{r} \biggl|1-\cos\left[\mathcal{R}_r\left(\mathbf{U}^{(c)}\left(I_r\right)\right),\mathbf{U}^{(c)}\left(I\right)\right]\biggl|
\end{aligned}
\end{equation}
To accelerate optimization, we schedule image downsampling according to image frequency \cite{chen2025dashgaussian}.
Given the sub-pixel density of the initialized Gaussians, we set the initial downsampling rate to 2 and apply only opacity and contribution-based pruning to eliminate low contributing primitives, together with Abs-based densification \cite{ye2024absgs}.

%% file: sec/4_exp.tex
\begin{figure*}[htpb]
\centering
\includegraphics[width = 0.99\textwidth]{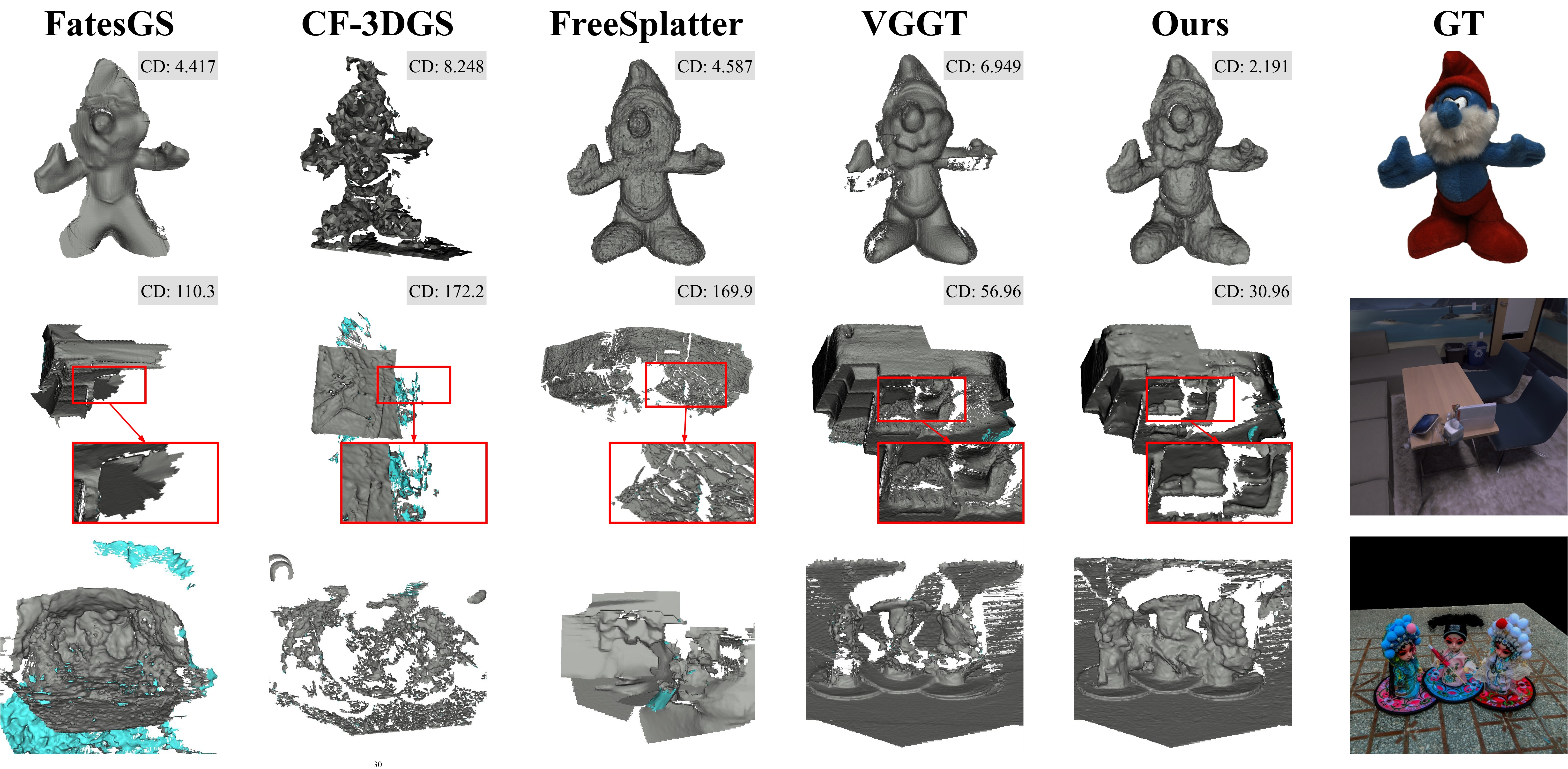}
\caption{Visualization of surface reconstruction on DTU and Replica datasets.}\label{fig_vis_mesh}
\end{figure*}

\begin{table*}[!ht]
\centering
\renewcommand{\arraystretch}{1.15}
\resizebox{\textwidth}{!}{%
\begin{tabularx}{1.6\textwidth}{c*{17}{>{\centering\arraybackslash}X}}
\hline
\textbf{Scan}          & \textbf{24}    & \textbf{37}    & \textbf{40}    & \textbf{55}    & \textbf{63}    & \textbf{65}    & \textbf{69}    & \textbf{83}    & \textbf{97}    & \textbf{105}   & \textbf{106}   & \textbf{110}   & \textbf{114}   & \textbf{118}   & \textbf{122}   & \textbf{Mean}  \\ \hline
3DGS-GT                & 3.013          & 4.935          & 5.461          & 1.398          & 3.743          & 3.033          & 1.892          & 4.854          & 3.360          & 1.899          & 2.469          & 3.431          & 1.068          & 1.780          & 1.850          & 2.946          \\
3DGS*-GT               & 2.010          & 4.382          & 5.277          & 1.873          & 5.771          & 3.885          & 2.035          & 4.676          & 4.197          & 1.819          & 5.241          & 3.914          & 1.507          & 5.866          & 3.613          & 3.738          \\
2DGS-GT                & 5.654          & 7.908          & 7.011          & 7.401          & 8.661          & 5.950          & 4.638          & 9.286          & 5.231          & 5.641          & 7.449          & 5.809          & 3.290          & 4.765          & 6.797          & 6.366          \\
2DGS*-GT               & 2.106          & 3.295          & 2.516          & 1.343          & 3.345          & 2.511          & 2.006          & 2.307          & 3.038          & 2.329          & 2.613          & 4.177          & 0.903          & 2.615          & 1.822          & 2.462          \\
PGSR-GT                & 3.087          & 3.577          & 3.001          & 1.032          & 3.390          & 2.482          & 2.294          & 2.925          & 3.815          & 1.704          & 2.419          & 3.410          & 1.057          & 1.928          & 1.968          & 2.539          \\
PGSR*-GT               & 3.360          & 3.280          & 2.750          & 1.270          & 5.150          & 1.840          & 0.880          & 1.790          & 3.490          & 1.190          & 1.860          & 1.110          & 0.610          & 1.090          & 1.520          & 2.079          \\
FatesGS-GT             & 1.058          & 3.615          & 5.159          & 0.888          & 1.509          & 2.025          & 0.890          & 3.404          & 1.578          & 1.235          & 1.389          & 0.814          & 0.498          & 0.909          & 0.966          & 1.729          \\
FatesGS*-GT            & 0.670          & 1.940          & 1.170          & 0.770          & 1.280          & 1.230          & 0.630          & 1.050          & 0.980          & 0.690          & 0.750          & 0.480          & 0.410          & 0.780          & 0.900          & 0.915          \\ \hdashline
3DGS                   & 6.999          & 9.218          & 8.181          & 6.916          & 10.86          & 6.262          & 5.188          & 10.26          & 5.483          & 5.447          & 7.216          & 6.438          & 3.969          & 5.242          & 6.725          & 6.961          \\
3DGS*                  & 5.759          & 9.947          & 8.383          & 7.127          & 9.431          & 6.293          & 5.480          & 8.755          & 6.106          & 6.507          & 8.211          & 7.454          & 4.245          & 6.184          & 7.421          & 7.154          \\
2DGS                   & 5.904          & 6.678          & 7.011          & 8.031          & 8.701          & 6.199          & 5.153          & 8.879          & 6.606          & 6.223          & 8.042          & 6.862          & 3.907          & 5.965          & 7.175          & 6.756          \\
2DGS*                  & 5.085          & 7.295          & 6.859          & 7.757          & 9.238          & 6.924          & 5.165          & 8.930          & 5.880          & 6.704          & 7.956          & 5.668          & 4.301          & 5.426          & 7.642          & 6.722          \\
PGSR                   & 5.359          & 6.676          & 6.628          & 7.168          & 7.922          & 5.913          & 4.733          & 8.571          & 5.532          & 5.646          & 7.188          & 4.797          & 3.037          & 4.569          & 6.534          & 6.018          \\
PGSR*                  & 3.745          & 8.030          & 6.021          & 7.004          & 8.612          & 6.006          & 4.574          & 9.085          & 4.618          & 6.163          & 7.511          & 3.879          & 3.226          & 4.687          & 6.718          & 5.992          \\
FatesGS                & 2.506          & 7.510          & 4.145          & 2.222          & 4.233          & 4.949          & 3.179          & 4.417 & 6.048          & 2.532          & 3.806          & 3.507          & 2.901          & 2.594          & 2.805          & 3.824          \\
FatesGS*               & 3.039          & 7.317          & 6.952          & 1.915          & 5.534          & 5.071          & 4.304          & \textit{4.237}    & 4.332          & 2.370          & 3.041          & 2.607          & 2.220          & 1.981          & 2.919          & 3.856          \\
CF-3DGS                & 8.488          & 9.307          & 8.346          & 9.740          & 8.800          & 8.132          & 6.818          & 8.248          & 8.919          & 7.755          & 8.442          & 7.437          & 7.841          & 7.860          & 8.009          & 8.276          \\
FreeSplatter           & 5.673          & 8.894          & 9.206          & 5.903          & 7.602          & 9.477          & 4.660          & 4.587          & 8.468          & 3.274          & 4.352          & 7.873          & 3.436          & 3.867          & 4.461          & 6.116          \\
MAtCha               & 1.461 	&6.528 	&2.324 	&1.126 	&3.908 	&2.375 	&1.547 	&3.144 	&2.509 	&1.625 	&1.716 	&2.676 	&0.950 	&1.562 	&1.723 	&2.345            \\
MAtCha*               & 1.393 	&5.102 	&2.261 	&1.285 	&4.680 	&2.843 	&1.773 	&4.703 	&3.284 	&2.964 	&2.368 	&3.115 	&0.910 	&1.873 	&2.961 	&2.768             \\
NoPoSplat             & 3.977 	&3.081 	&4.728 	&3.062 	&6.415 	&6.518 	&6.702 	&6.273 	&7.333 	&3.903 	&5.200 	&7.894 	&2.028 	&6.787 	&6.491 	&5.359              \\
MVSplat             & 8.748 	&7.941 	&5.057 	&6.720 	&9.212 	&6.791 	&7.936 	&7.489 	&6.702 	&9.873 	&9.289 	&6.269 	&10.86 	&6.083 	&7.010 	&7.732 \\
VGGT                   & \textit{2.417} & \textit{4.967} & \textit{2.225} & {\ul 1.806}    & \textit{3.507} & \textit{2.315} & \textit{1.681} & 6.949          & \textit{2.362} & \textit{2.012} & \textit{1.863} & {\ul 1.845}    & \textit{1.044} & \textit{2.030} & \textit{1.761} & \textit{2.586} \\
\textbf{Ours(wo Opt.)} & {\ul 1.592}    & {\ul 2.580}    & {\ul 2.040}    & \textit{1.878} & {\ul 2.950}    & {\ul 1.743}    & {\ul 1.572}    & {\ul 2.303}          & {\ul 2.167}    & {\ul 1.534}    & {\ul 1.689}    & \textit{2.052} & {\ul 0.921}    & {\ul 1.895}    & {\ul 1.692}    & {\ul 1.907}    \\
\textbf{Ours}          & \textbf{1.300} & \textbf{2.103} & \textbf{1.767} & \textbf{1.468} & \textbf{2.066} & \textbf{1.476} & \textbf{1.339} & \textbf{2.191} & \textbf{1.736} & \textbf{1.479} & \textbf{1.419} & \textbf{1.420} & \textbf{0.793} & \textbf{1.572} & \textbf{1.583} & \textbf{1.581} \\ \hline

\end{tabularx}
}
\caption{Quantitative experiments of surface reconstruction CD$\downarrow$ (mm) on DTU datasets under various settings.}\label{tbl-dtucd-1}
\end{table*}

\section{Experiments}
\subsection{Experimental settings}
\paragraph{Training}The model was trained with a $512\times512$ resolution and a patch size of 2048. Our experiments are conducted on an Intel 8470Q CPU and an Nvidia RTX5090 GPU, requiring 27 hours to complete 120 epochs.
The training dataset consists of BlendMVS \cite{yao2020blendedmvs}, DTU \cite{dtu}, Virtual KITTI \cite{Vkitti2}, and Replica \cite{replica19arxiv}, encompassing extensive RGB-D images and camera parameters from both synthetic and real-captured sources.

\vspace{-5mm}
\paragraph{Metrics}We employ widely adopted metrics including PSNR, SSIM \cite{SSIM}, and LPIPS \cite{LPIPS} to evaluate NVS quality; employ Chamfer Distance (CD) \cite{dtu} to evaluate surface reconstruction quality; employ RMSE \cite{gsicpslam} (rotation and translation) to evaluate camera pose estimation. 
\vspace{-5mm}

\begin{table*}[!ht]
\centering
\renewcommand{\arraystretch}{1.0}
\resizebox{\textwidth}{!}{%
\begin{tabular}{p{1.5cm}ccccc|ccccccccc}

\hline
                             &       & \makecell[c]{3DGS\\ GT} & \makecell[c]{2DGS\\ GT} & \makecell[c]{PGSR\\ GT} & \multicolumn{1}{c|}{\makecell[c]{FatesGS\\ GT}} & 3DGS           & 2DGS           & PGSR  & FatesGS & \makecell[c]{CF\\ 3DGS} & \makecell{Free\\Splatter} & VGGT  & \makecell{\textbf{Ours}\\ \textbf{(wo Opt.)}} & \textbf{Ours}  \\ \hline
\multicolumn{1}{c|}{}        & PSNR$\uparrow$  & 34.35                                             & 28.00                                             & 31.01                                             & 23.00                                                & {{\ul 29.37}}    & 26.58          & 27.13 & 23.76   & 25.97                                             & 18.26        & 18.36 & \textit{28.61}         & \textbf{30.251} \\
\multicolumn{1}{c|}{DTU}     & SSIM$\uparrow$  & 0.933                                             & 0.873                                             & 0.937                                             & 0.801                                                & {\textit{0.878}} & 0.858          & 0.872 & 0.819   & 0.856                                             & 0.655        & 0.771 & {\ul 0.890}            & \textbf{0.909} \\
\multicolumn{1}{c|}{}        & LPIPS$\downarrow$ & 0.153                                             & 0.215                                             & 0.111                                             & 0.336                                                & {\textit{0.210}} & 0.230          & 0.185 & 0.250   & 0.167                                             & 0.244        & 0.194 & {\ul 0.132}            & \textbf{0.110} \\ \hline
\multicolumn{1}{c|}{}        & PSNR$\uparrow$  & 22.40                                             & 21.35                                             & 22.07                                             & 22.28                                                & {\textit{22.05}} & 21.27          & 21.44 & 22.53   & 18.34                                             & 14.55        & 13.04 & {\ul 23.51}            & \textbf{35.79} \\
\multicolumn{1}{c|}{Replica} & SSIM$\uparrow$  & 0.804                                             & 0.800                                             & 0.773                                             & 0.806                                                & {{\ul 0.802}}    & \textit{0.799} & 0.767 & 0.808   & 0.695                                             & 0.628        & 0.531 & 0.758                  & \textbf{0.965} \\
\multicolumn{1}{c|}{}        & LPIPS$\downarrow$ & 0.409                                             & 0.416                                             & 0.418                                             & 0.402                                                & {0.411}          & 0.412          & 0.428 & 0.402   & \textit{0.387}                                    & 0.458        & 0.514 & {\ul 0.382}            & \textbf{0.095} \\ \hline
\multicolumn{1}{c|}{}           & PSNR$\uparrow$    & 17.73                                                                   & 17.24                                                                   & 17.63                                                                   & 16.56                                                                      & \textit{18.75} & 17.81         & 18.65 & 18.31         & 12.93                                                                   & 15.18                                                               & 13.41         & {\ul 19.52}                                                                & \textbf{30.74} \\
\multicolumn{1}{c|}{BlendedMVS} & SSIM$\uparrow$    & 0.547                                                                    & 0.543                                                                    & 0.544                                                                    & 0.535                                                                       & 0.565           & 0.549          & 0.563  & \textit{0.565} & 0.432                                                                    & 0.488                                                                & 0.527          & {\ul 0.634}                                                                 & \textbf{0.965}  \\
\multicolumn{1}{c|}{}           & LPIPS$\downarrow$ & 0.499                                                                    & 0.501                                                                    & 0.490                                                                    & 0.506                                                                       & 0.483           & 0.495          & 0.475  & 0.477          & 0.453                                                                    & 0.491                                                                & \textit{0.411} & {\ul 0.353}                                                                 & \textbf{0.052}  \\ \hline

\end{tabular}
}
\caption{Quantitative evaluation of NVS on DTU, Replica, and BlendedMVS datasets}\label{tbl-psnr-1}
\vspace{-5mm}
\end{table*}



\paragraph{Baselines}We evaluate the reconstruction results with \textit{base} methods: 3DGS \cite{kerbl3DGS}; 2DGS \cite{2dgs_Huang}.
\textit{Pose-free} methods: CF-3DGS \cite{Fu_2024_CVPR_cf3dgs}; FreeSplatter \cite{xu2024freesplatter}; VGGT \cite{wang2025vggt}; NoPoSplat \cite{ye2024noposplat}; MAtCha \cite{guedon2025matcha}. \textit{Sparse-view surface reconstruction} methods: FatesGS \cite{huang2025fatesgs}; PGSR \cite{chen2024pgsr}; MVSplat \cite{chen2024mvsplat}.

3DGS, 2DGS, FatesGS, and PSGR, as classical GS-based per-scene optimization pipelines, require camera parameters as input. 
We obtain camera parameter priors from VGGT and evaluate the performance of these methods under both ground truth and estimated camera parameters. 
Furthermore, we conduct comparative evaluations under equal iterations (1000) and default iterations.

\vspace{-5mm}

\paragraph{Evaluation}We evaluate FSFSplatter on DTU \cite{dtu}, Replica \cite{replica19arxiv} and BlendedMVS \cite{yao2020blendedmvs} datasets. DTU datasets have been extensively used in surface reconstruction. We evaluate on all 15 scenes, where each scene contains 49 or 64 images at a resolution of $1600 \times 1200$. Following previous works \cite{huang2025fatesgs, yu2021pixelnerf}, we select views 23, 24, and 33 as inputs and downsample the images to $800 \times 600$. Since DTU datasets primarily consist of object-level scenes, we further validate on Replica and BlendedMVS datasets to assess generalization capability and performance on scene-level scenes.
We respectively select the 5 scenes and randomly choose 3 input views at resolutions of $1200 \times 680$ and $768 \times 576$. 



\subsection{Evaluation of Surface Reconstruction}
\hspace{1em}We first evaluate the surface reconstruction accuracy (CD) under different settings across all baselines. As shown in \cref{tbl-dtucd-1}, \cref{tbl-replicacd-1}, and \cref{fig_vis_mesh}, FSFSplatter demonstrates significant advantages in CD error even when compared with per-scene optimization methods using their default iteration numbers, and remains competitive with methods using ground truth camera parameters as input. Particularly in scene-level scenarios, FSFSplatter exhibits superior robustness. The error is reduced by at least 28.39\% and 34.37\% on DTU and Replica datasets respectively. Even without per-scene optimization, the surface reconstruction error is reduced by at least 14.62\% and 21.93\% respectively.

In the tables, GT indicates the use of ground truth camera parameters as input, and * indicates the use of official default iteration numbers. \textbf{Bold}, {\ul underline}, and \textit{italic} text represent the top three accuracy rankings respectively.

\begin{table}[!ht]
\centering
\renewcommand{\arraystretch}{1.0}
\resizebox{0.5\textwidth}{!}{%
\begin{tabularx}{0.7\textwidth}{w{c}{1.3cm}*{9}{>{\centering\arraybackslash}X}}
\hline
\textbf{Scan}          & \textbf{Office0} & \textbf{Office1} & \textbf{Office2} & \textbf{Office3}& \textbf{Office4} & \textbf{Room0}  & \textbf{Room1} & \textbf{Room2}  & \textbf{Mean}   \\ \hline
3DGS-GT & 183.52 & 179.71 & 172.41 & 170.82 & 183.96 & 177.93 & 176.67 & 186.65 & 178.96 \\
3DGS*-GT & 177.36 & 163.13 & 176.42 & 174.87 & 146.27 & 167.02 & 173.80 & 144.06 & 165.37 \\
2DGS-GT & 179.50 & 179.90 & 166.42 & 179.11 & 181.54 & 183.91 & 166.79 & 157.23 & 174.30 \\
2DGS*-GT & 152.44 & 144.69 & 187.64 & 168.32 & 179.59 & 181.86 & 168.51 & 189.26 & 171.54 \\
PGSR-GT & 168.11 & 183.27 & 167.57 & 185.56 & 185.53 & 181.68 & 174.74 & 182.66 & 178.64 \\
PGSR*-GT & 170.03 & 171.97 & 177.46 & 178.15 & 164.49 & 176.56 & 174.64 & 158.07 & 171.42 \\
FatesGS-GT & 181.22 & 176.32 & 183.77 & 189.85 & 181.15 & 178.73 & 188.13 & 179.21 & 182.30 \\
FatesGS*-GT & 187.27 & 153.78 & 183.85 & 166.94 & 112.96 & 186.56 & 175.29 & 187.77 & 169.30 \\ \hdashline
3DGS & 167.24 & 163.32 & 177.39 & 174.28 & 159.15 & 197.04 & 178.15 & 174.62 & 173.90 \\
3DGS* & 176.50 & 164.00 & 177.84 & 192.52 & 140.18 & 175.12 & 158.42 & 140.58 & 165.65 \\
2DGS & 181.87 & 149.37 & 196.53 & 179.11 & 181.52 & 176.93 & 165.06 & 157.21 & 173.45 \\
2DGS* & 176.00 & 162.06 & 172.93 & 168.30 & 179.56 & 174.21 & 161.86 & 189.26 & 173.02 \\
PGSR & 166.96 & 185.52 & 177.85 & 187.53 & 139.26 & 191.64 & 166.48 & 153.03 & 171.03 \\
PGSR* & 147.42 & 140.70 & 185.09 & 159.03 & 155.95 & 192.01 & 170.16 & 150.38 & 162.59 \\
FatesGS & 160.64 & 57.146 & 189.05 & 162.95 & 153.01 & 161.51 & 171.60 & 158.90 & 151.85 \\
FatesGS* & 110.32 & 95.371 & 183.05 & 165.24 & 161.20 & 181.56 & 153.97 & 154.94 & 150.71 \\
CF-3DGS & 172.16 & 165.08 & 152.55 & 179.37 & 163.13 & 173.52 & 184.27 & 181.09 & 171.40 \\
FreeSplatter & 169.90 & 158.22 & 172.16 & 181.14 & 142.32 & 163.02 & 173.96 & 160.15 & 165.11 \\
VGGT & \textit{56.961} & {\ul 35.937} & \textit{77.528} & \textit{83.141} & \textit{36.031} & \textit{43.357} & \textit{42.706} & {\ul 30.885} & \textit{50.818} \\
\textbf{Ours(wo Opt.)} & {\ul 39.047} & \textit{41.558} & {\ul 66.043} & \textbf{33.762} & {\ul 33.248} & \textbf{26.298} & {\ul 27.282} & \textit{33.729} & {\ul 37.621} \\
\textbf{Ours} & \textbf{30.962} & \textbf{33.914} & \textbf{51.489} & {\ul 39.033} & \textbf{25.041} & {\ul 28.079} & \textbf{23.866} & \textbf{28.050} & \textbf{32.554} \\ \hline
\end{tabularx}
}

\caption{Quantitative experiments of surface reconstruction CD$\downarrow$ (mm) on Replica datasets under different settings. We conduct experiments on 5 scenes.}\label{tbl-replicacd-1}
\vspace{-4mm}
\end{table}

\begin{figure}[htpb]
\centering
\includegraphics[width = 0.49\textwidth]{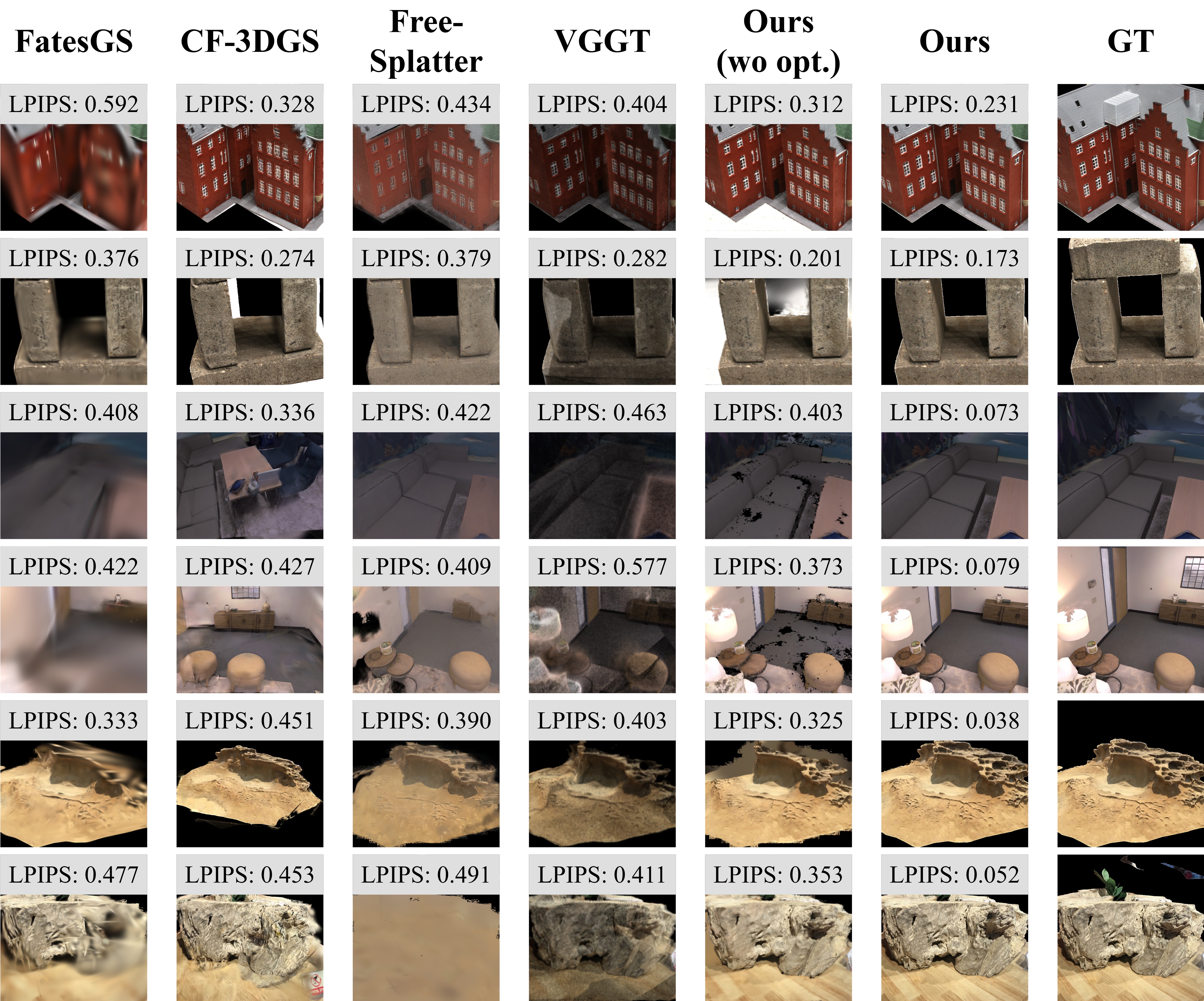}
\vspace{-4mm}
\caption{Visualization of NVS on DTU, Replica, and BlendedMVS datasets.}\label{fig_vis_nvs}
\vspace{-6mm}
\end{figure}

\subsection{Evaluation of NVS}

\hspace{1em}Subsequently, we evaluate the performance of FSFSplatter on NVS. As demonstrated in \cref{tbl-dtucd-1} and \cref{tbl-replicacd-1}, under sparse views, simply increasing the number of iterations does not yield better reconstruction; conversely, it leads to catastrophic scene geometry due to overfitting. Therefore, we evaluate NVS performances under equal iterations. As shown in \cref{tbl-psnr-1} and \cref{fig_vis_nvs}, the NVS performance of FSFSplatter significantly surpasses other approaches. Even without per-scene optimization,
FSFSplatter still demonstrates a clear advantage.
Moreover,
our approach effectively preserves scene geometry while performing per-scene optimization. The NVS error (LPIPS as an example) is reduced by at least 46.19\% on DTU datasets, 73.13\% on Replica datasets, and 87.35\% on BlendedMVS datasets. Even without per-scene optimization, the error is decreased by at least 37.14\%, 1.29\%, and 14.36\% respectively.

\subsection{Evaluation of pose estimation}
\hspace{1em}As a surface reconstruction and NVS method based on free sparse-view images, we also evaluated the accuracy of camera pose estimation on Replica and DTU datasets.
As shown in \cref{tbl-pose-1}, FSFSplatter achieves competitive performance in terms of RMSE \cite{gsicpslam} compared to SOTA pose estimation methods, while simultaneously delivering high-quality surface reconstruction and NVS results.

\begin{table}[!ht]
\centering
\renewcommand{\arraystretch}{1.6}
\resizebox{0.5\textwidth}{!}{%
\begin{tabularx}{0.73\textwidth}{w{c}{1.0cm}c*{6}{>{\centering\arraybackslash}X}}

\hline
\textbf{}                &                            & \makecell[c]{CF\\3DGS} & \makecell[c]{Free\\ Splatter} & Fast3R & Dust3r & VGGT  & \textbf{Ours}  \\ \hline
\multirow{2}{*}{Replica} & \multicolumn{1}{c|}{Rot($^\circ$)$\downarrow$}   & 45.267           & 47.136                & \textit{3.432}  & 39.895          & {\ul 0.647}    & \textbf{0.634} \\
                         & \multicolumn{1}{c|}{Trans(mm)$\downarrow$} & 135.50          & 107.64               & \textbf{3.637}  & 7.806           & \textit{3.837} & {\ul 3.746}    \\\hdashline
\multirow{2}{*}{DTU}     & \multicolumn{1}{c|}{Rot($^\circ$)$\downarrow$}   & 5.909            & 38.271                & \textit{2.581}  & 42.939          & \textbf{0.301} & {\ul 0.314}    \\
                         & \multicolumn{1}{c|}{Trans(mm)$\downarrow$} & 20.762           & 70.912                & 14.212          & \textit{10.519} & {\ul 1.299}    & \textbf{1.256} \\ \hline

\end{tabularx}
}
\caption{RMSE on Replica and DTU datasets.}\label{tbl-pose-1}
\end{table}

\vspace{-5mm}
\subsection{Evaluation of optimization speed}
\hspace{1em}As a reconstruction method that supports per-scene optimization, FSFSplatter achieves high reconstruction accuracy from the initial stage due to well-geometry Gaussian dense initialization. Consequently, FSFSplatter attains remarkably short optimization iterations. As shown in \cref{tbl-time-1}, we evaluate the optimization speed of FSFSplatter and other per-scene reconstruction algorithms on Replica datasets.

\begin{table}[!ht]
\centering
\renewcommand{\arraystretch}{1.0}
\resizebox{0.4\textwidth}{!}{%
\begin{tabularx}{0.4\textwidth}{*{2}{>{\centering\arraybackslash}X}}
\hline
\textbf{Method} & \textbf{Time}$\downarrow$(s) \\ \hline
3DGS            & {\ul 429.45}        \\
2DGS            & \textit{840.37}        \\
FatesGS         & 915.93        \\
PGSR            & 1018.2        \\
Ours            & \textbf{107.39}        \\ \hline
\end{tabularx}
}
\caption{Optimization speed evaluation of per-scene optimization-based algorithms on Replica datasets.}\label{tbl-time-1}
\end{table}

\vspace{-3mm}
On DTU dataset, FSFSplatter's initialization requires 0.63s, which increases  \textless 0.27s over the baseline (VGGT \cite{wang2025vggt}: 0.24s + GS init \cite{kerbl3DGS}: 0.13s), and yields a 26.3\% improvement in surface reconstruction, an 11.9dB PSNR increase in NVS, and an 80\% reduction in subsequent optimization time.
Compared to pointcloud generators (Dust3R \cite{dust3rwang}: 6.4s, MASt3R \cite{mast3r_eccv24}: 4.4s, Fast3R \cite{Yang_2025_Fast3R}: 0.004s) and GS generators (NoPoSplat \cite{ye2024noposplat}: 4.16s, MVSplat \cite{chen2024mvsplat}: 0.79s, FreeSplatter \cite{xu2024freesplatter}: 4.29s), our approach maintains competitive efficiency.

\subsection{Ablation study}
\hspace{1em}To validate the effectiveness of each component in FSFSplatter, we conducted comprehensive ablation studies. Specifically, we performed ablations on both end-to-end dense Gaussian initialization and geometrically-enhanced scene optimization, with the surface reconstruction quality evaluated using CD. Experiments were carried out on DTU and Replica datasets.
For per-scene optimization, we ablated the following components: monocular depth supervision $\mathcal{L}_{\text{rank}}$, depth smoothness loss $\mathcal{L}_\text{smooth}$, multi-view feature consistency supervision $\mathcal{L}_{\text{MVS}}$, and differentiable camera poses $T^\text{cam}_k$. For Gaussian initialization, we ablated the contribution-based Gaussian pruning $\mathbb{T}$ and the self-splitting-based Gaussian scene densification $\mathbb{D}$.

\begin{table}[!ht]
\centering
\renewcommand{\arraystretch}{1}
\resizebox{0.5\textwidth}{!}{%
\begin{tabularx}{0.6\textwidth}{c*{3}{>{\centering\arraybackslash}X}}
\hline
              & \multicolumn{2}{c}{CD$\downarrow$} \\ \cline{2-3} 
              & Replica     & DTU      \\ \hline
\textbf{Ours}         & 33.66       & 1.581    \\
No $\mathcal{L}_{\text{rank}}$     & 36.23       & 2.905    \\
No $\mathcal{L}_{\text{smooth}}$     & 37.71       & 2.653    \\
No $\mathcal{L}_{\text{MVS}}$      & 35.36       & 3.130    \\
No $T^\text{cam}_k$ & 37.05       & 2.855    \\
\hdashline
\textbf{Ours(wo Opt.)} & 40.05       & 2.208    \\
No $\mathbb{T}$      & 40.69       & 2.930    \\
No $\mathbb{D}$   & 59.83       & 3.584    \\ \hline
\end{tabularx}
}
\vspace{-1mm}
\caption{Ablation experiments on DTU and Replica datasets.}\label{tbl-abla-1}
\vspace{-3mm}
\end{table}

As shown in \cref{tbl-abla-1}, the removal of any individual component leads to a degradation in the surface reconstruction quality, thereby demonstrating the contribution of each component. It is worth noting that in the per-scene optimization stage, removing additional supervision may result in a reconstructed surface that is inferior to the initial scene. This can be primarily attributed to the geometric correctness of the dense Gaussian initialization and the inherent ambiguity in sparse viewpoints, which prevent the scene geometry enhancement from simple increases in iteration counts.

%% file: sec/5_conclusion.tex
\section{Conclusion}
\hspace{1em}We present FSFSplatter, a fast and geometrically accurate method for free sparse-view reconstruction. Our end-to-end transformer jointly performs dense Gaussian initialization and camera estimation, with a self-splitting Gaussian head for dense scene generation and contribution-based pruning for floater removal.
To alleviate sparse-view overfitting caused by sparse views, we enforce monocular depth supervision, multi-view feature consistency, and differentiable camera poses from camera parameter estimation. FSFSplatter outperforms prior methods in both accuracy and robustness, with and without per-scene optimization, and achieves SOTA results for surface reconstruction and novel view synthesis on widely used DTU, Replica, and BlendedMVS.
In future work, we plan to integrate more prior information, such as depth maps and known intrinsics, into the pipeline as a potential direction.

%% file: sec/6_Acknowledgments.tex
\section*{Acknowledgments}
The authors gratefully acknowledge the financial support from the Shanghai Science and Technology Action Plan (Grant No. 21JM0010300), the Shanghai Aerospace Science and Technology Innovation Fund (SAST) (Grant No. 2021-037), the Special Fund for Technology Innovation Support Projects of Shanghai (Grant Nos. 2021-cyxt-kj1, XTCX-KJ-2022-37, and HCXBCY-2023-046), the National Defense Basic Scientific Research Program of China(Grant No. JCKY2021606B002), and the Zhejiang Provincial “Pioneer \& Leading Goose + X” Science and Technology Program (Grant No. 2026C02A1220).